\documentclass{article}
\usepackage{spconf,amsmath,graphicx}
\usepackage{subfigure}
\usepackage{array}
\usepackage{booktabs}
\usepackage{multirow}
\usepackage{lipsum}

\newcommand\blfootnote[1]{%
  \begingroup
  \renewcommand\thefootnote{}\footnote{#1}%
  \addtocounter{footnote}{-1}%
  \endgroup
}

\graphicspath{{./images/}}

\title{The importance of stain normalization in colorectal tissue classification with convolutional networks}

\makeatletter
\def\@name{ \emph{Francesco Ciompi$^{1,2}$},  \emph{Oscar Geessink$^{1,2,3}$}, \emph{Babak Ehteshami Bejnordi$^{1,2}$}, \emph{Gabriel Silva de Souza$^{1}$}\\
\emph{Alexi Baidoshvili$^{3}$}, \emph{Geert Litjens$^{1,2}$}, \emph{Bram van Ginneken$^{2}$}, \emph{Iris Nagtegaal$^{1}$}, \emph{Jeroen van der Laak$^{1,2}$}}
\makeatother

\address{\small{$^{1}$ Dept. of Pathology, Radboud University Medical Center, Nijmegen, Netherlands} \\ \small{$^{2}$Diagnostic Image Analysis Group, Dept. of Radiology, Radboud University Medical Center, Nijmegen, Netherlands} \\ \small{$^{3}$ Laboratorium Pathologie Oost Nederland, Hengelo, Netherlands}}
\begin{document}
%
\maketitle
\begin{abstract}
The development of reliable imaging biomarkers for the analysis of colorectal cancer (CRC) in hematoxylin and eosin (H\&E) stained histopathology images requires an accurate and reproducible classification of the main tissue components in the image.
In this paper, we propose a system for CRC tissue classification based on convolutional networks (ConvNets).
We investigate the importance of stain normalization in tissue classification of CRC tissue samples in H\&E-stained images.
Furthermore, we report the performance of ConvNets on a cohort of rectal cancer samples and on an independent publicly available dataset of colorectal H\&E images.
\end{abstract}
%

\section{Introduction}
\label{sec:intro}

Adjuvant chemotherapy has been shown to significantly increase survival for some patients affected by colorectal cancer (CRC), a disease that has a high global incidence of over 1.3 million cases per year, causing 694,000 deaths annually\footnote{globocan.iarc.fr/Pages/fact\_sheets\_cancer.aspx}.
The decision whether a patient should be treated with adjuvant chemotherapy depends mostly on the tumor stage.
However, patients diagnosed with the same stage of disease can have considerable diverse outcomes.
This indicates the need for additional biomarkers, beside tumor stage, that will allow further stratification of CRC patients and identify those who may or may not benefit from adjuvant treatment. 

In recent years, researchers have been investigating the role of histological parameters as new biomarkers to guide adjuvant treatment decision.
Examples are the proportion of necrosis \cite{Poll10} and the relative amounts of tumor and stroma \cite{Huij13}, which proved to be strong independent prognostic factors in CRC.
Despite the great potential of these new biomarkers, their manual assessment suffers from limited clinical applicability and high inter- and intra-observer variability.
In this context, the automatic analysis of digitized whole-slide histopathology images (WSI) can contribute to the development of objective and reproducible imaging biomarkers for the field of clinical pathology.

A key element in the design of imaging biomarkers based on WSI analysis is the automated pixel-wise classification of relevant tissues and histological structures.
In recent years, convolutional networks (ConvNets) \cite{Lecu15} have become the reference algorithm to solve image- and patch-based classification tasks in medical imaging.
Recent challenges in digital pathology\footnote{warwick.ac.uk/fac/sci/dcs/research/combi/research/bic/glascontest/}\textsuperscript{,}\footnote{camelyon16.grand-challenge.org/}\textsuperscript{,}\footnote{tupac.tue-image.nl/} showed that methods based on ConvNets can perform as well or better than pathologists at analyzing hematoxylin and eosin (H\&E) stained histopathology images.

\begin{figure}[t]
\centering
{\includegraphics[width=0.19\linewidth]{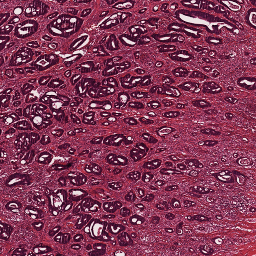}}\hspace{0.0em}
{\includegraphics[width=0.19\linewidth]{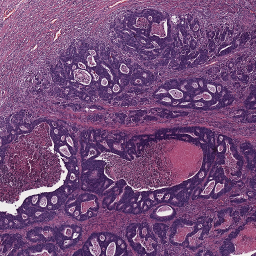}}\hspace{0.0em}
{\includegraphics[width=0.19\linewidth]{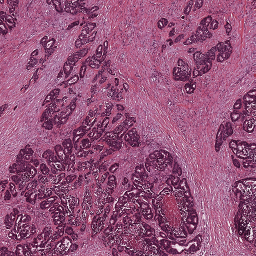}}\hspace{0.0em}
{\includegraphics[width=0.19\linewidth]{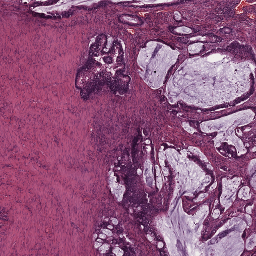}}\hspace{0.0em}
{\includegraphics[width=0.19\linewidth]{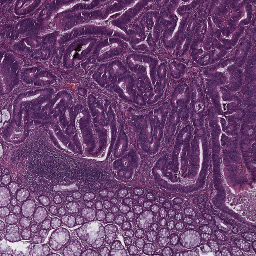}}\hspace{0.0em}\\
\vspace{0.15cm}
{\includegraphics[width=0.19\linewidth]{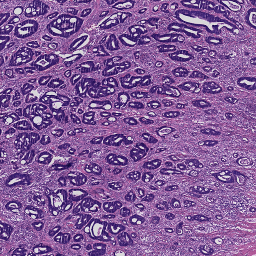}}\hspace{0.0em}
{\includegraphics[width=0.19\linewidth]{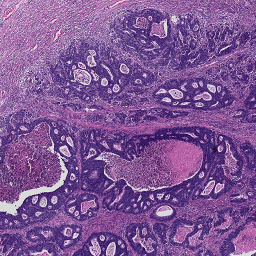}}\hspace{0.0em}
{\includegraphics[width=0.19\linewidth]{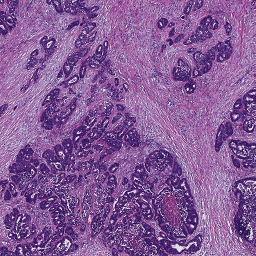}}\hspace{0.0em}
{\includegraphics[width=0.19\linewidth]{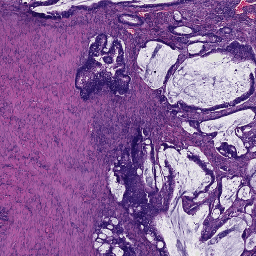}}\hspace{0.0em}
{\includegraphics[width=0.19\linewidth]{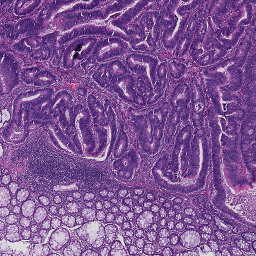}}\hspace{0.0em}\\
\caption{Samples of CRC tissue from \cite{Kath16} (top) and their stain-normalized version (bottom) using the algorithm in \cite{Ehte16}.\vspace{-0.3cm}
\label{fig:stain_normalization}}
\end{figure}

\begin{table*}[t]
\scriptsize
\renewcommand{\arraystretch}{1.3}
\centering
\begin{tabular}{| c |  c | c | c | c | c | c | c | c | c | c |}
\hline
\multicolumn{11}{|c|}{7 weight layers}\\
\hline \hline
conv5-32 & MP & conv5-64 & MP & conv3-128 & MP & conv3-256 & MP & conv9-1024 & conv1-512 & SM\\
\hline
\end{tabular}
\caption{ConvNets architecture. The nomenclature follows the one used in \cite{Simo14}. MP = max-pooling layer, SM = soft-max layer.}
\label{tab:convnet}
\end{table*}

The procedure of fixation, embedding, cutting and staining of tissue sections affects the appearance of H\&E stained histology samples, which can vary significantly across laboratories, but even across staining batches within the same lab. 
Although this variability only partially limits the interpretation of images by pathologists, it can dramatically affect the result of automatic image analysis algorithms.
To cope with this problem, stain normalization (SN) algorithms for histological images have been recently developed \cite{Ehte16,Mace09,Khan14a}, with the aim of matching stain colors of WSI with a given template.
These SN algorithms have shown great promise to deal with stain variations.
However, the benefit of stain normalization in problems of patch-based tissue classification with ConvNets has never been investigated.

In this paper, we investigate the importance of stain normalization in tissue classification of H\&E-stained CRC tissue samples with convolutional networks.
Our contribution is three-fold.
First, we propose a convolutional network architecture to classify 9 tissue types of rectal cancer tissue samples, which we train and validate on a set of 74 whole-slide images.
Second, we investigate the applicability of the representation learned by ConvNets on rectal cancer data from one source to CRC data from different sources using a recent publicly available dataset of CRC data.
In this procedure, we investigate the role of two state of the art SN algorithms, comparing results with and without stain normalization.
Finally, we address the question how stain normalization should be used in a tissue classification pipeline.

\begin{figure}[t]
\centering
{\includegraphics[width=1\linewidth]{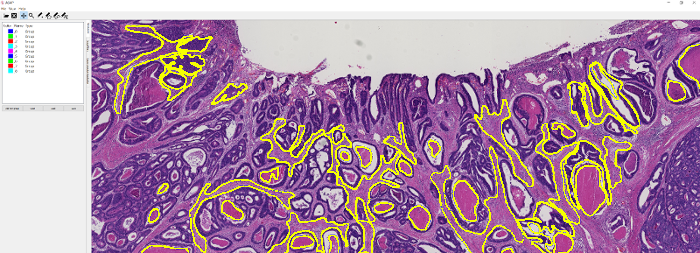}}\hspace{0.0em}
\caption{Example of manual annotations in a rectal cancer sample using the in-house developed open source software ASAP.\vspace{-0.3cm}
\label{fig:annotations}}
\end{figure}

\begin{table*}[t]
\scriptsize
\renewcommand{\arraystretch}{1.3}
\centering
\begin{tabular}{|c|c|c|c|c|c|c|c|c|c|}
\hline
\textbf{CRC classes} & Tumor epithelium 	& Simple + complex stroma 	& Immune cells & Debris and mucus & Mucosal glands & Adipose tissue & Background \\
\hline
\textbf{RC classes}   & Tumor 	& Stroma	 + muscle			& Lymphocytes	&	Necrosis + blood + mucus 	& Healthy epithelium	& Fatty tissue & - \\
\hline
\end{tabular}
\caption{Correspondence of classes in the CRC and the classes used in the RC dataset.}
\label{tab:classes}
\end{table*}

\section{Method}
In this section, we introduce (1) the data used to develop and validate the proposed method for CRC tissue classification, (2) the approach to build and train a convolutional network, and (3) the algorithms used to investigate the role of stain normalization in CRC tissue classification with ConvNets.\vspace{-0.3cm}

\subsection{Material}
In this paper, we used data from two different sources, namely a cohort of whole-slide images from rectal cancer samples, and a dataset of colorectal cancer images and patches, which was recently made publicly available \cite{Kath16}.
We used the cohort of rectal cancer WSIs to train convolutional networks and to validate the performance in cross-validation.
Successively, we used the CRC dataset to further validate the performance of ConvNets on an independent dataset and to investigate the role of stain normalization in CRC tissue classification.\vspace{-0.1cm}

\paragraph*{Rectal cancer data.}
A set of 74 histological slides from 74 patients was prepared from surgically excised rectal carcinomas in patients who had not received neoadjuvant chemotherapy and/or radiotherapy.
Slide preparation involved standard fixation of 2$\mu$m tissue sections and H\&E staining.
Whole-slide scanning was performed at 200X magnification (pixel resolution = 0.455 $\mu$m) using a Hamamatsu NanoZoomer 2.0-HT C9600-13 scanner (Herrsching, Germany), which produce 74 gigapixel whole-slide images.
Manual annotations of 9 tissue classes were made by an expert using the open source software ASAP\footnote{github.com/GeertLitjens/ASAP} developed in-house, which included: (i) tumor, (ii) stroma, (iii) necrosis, (iv) muscle, (v) healthy epithelium, (vi) fatty tissue, (vii) lymphocytes, (viii) mucus and (ix) blood cells (see Figure \ref{fig:annotations}).
All annotations were successively thoroughly checked by a pathologist and corrections were made when necessary.
In the rest of the paper, we refer to this dataset as rectal cancer (RC) data.\vspace{-0.3cm}

\paragraph*{Colorectal cancer data.}
A dataset of colorectal cancer images and patches from 10 patients was recently made publicly available \cite{Kath16}.
The test dataset consists of two subsets.
The first subset contains 5000 patches of 150$\times$150 pixel extracted from 10 H\&E slides of CRC cases.
The patches contain 625 examples of 8 tissue types, namely 
(i) tumor epithelium,
(ii) simple stroma,
(iii) complex stroma,
(iv) immune cell conglomerates, 
(v) debris and mucus,
(vi) mucosal glands,
(vii) adipose tissue,
(viii) background.
We name this subset as CRC$_p$.
The second subset contains 10 tiles of size 5000$\times$5000 pixel of H\&E stained CRC tissue samples, which we call CRC$_t$.\vspace{-0.1cm}

\subsection{Convolutional Network}
The design of ConvNets (see Table \ref{tab:convnet}) was based on the approach proposed in \cite{Simo14} and consists of a fully convolutional network \cite{Long15} with 11 layers.
The input of the network is an RGB patch of size 150$\times$150 pixels, which is the patch size in CRC$_p$.
As in \cite{Simo14}, we alternated convolutional (with ReLU non linearity) and max-pooling layers, starting with 32 filters in the first layer, and doubling the amount of filters after every max-pooling, but we used slightly larger filters (5$\times$5) in the first two layers.
The output of the network is a 9-elements vector containing the probability of the patch to belong to one of the 9 RC tissue types.

\subsection{Stain normalization}\label{sec:stain_normalization}
Stain normalization (SN) involves transforming an image $I$ into another image $\hat{I}$, through the operation $\hat{I} = f(I, \theta)$, where $\theta$ is a set of parameters extracted from a predefined template image and $f$ is the mapping function that matches the visual appearance of a given image to the template image.
The parameters $\theta$ are generally defined to capture color information of the main stain components in the image (e.g. H and E).
As a result, stain-normalized images will have a distribution of colors that resemble the ones of the template.
In Figure \ref{fig:stain_normalization}, images from CRC$_t$ and their stain-normalized version based on the WSI template of Figure \ref{fig:annotations} are depicted.

In this paper, we considered two state of the art SN algorithms.
The first one is based on the method recently published by Bejnordi et al. \cite{Ehte16}, which we refer to as SN$_1$.
The second one is based on the method published by Macenko et al. \cite{Mace09}, which was mentioned as the reference method for stain normalization in the TUPAC\footnote{tupac.tue-image.nl/} and AMIDA\footnote{amida13.isi.uu.nl/} challenges and has a publicly available MATLAB implementation\footnote{github.com/mitkovetta/staining-normalization}.
We refer to this method as SN$_2$.\vspace{-0.2cm}

\begin{table}[t]
\scriptsize
\renewcommand{\arraystretch}{1.3}
\centering
\begin{tabular}{|c|c|c|c|}
\hline
	& training w/ SN	& training w/o SN\\
\hline
testing w/ SN	& \textbf{79.66\%} [C]	& 75.55\% [B]\\
\hline
testing w/o SN	& 45.65\% [D]	& 50.96\% [A]\\
\hline
\end{tabular}
\caption{Accuracy with and without stain normalization (SN) in the training and testing procedure.}
\label{tab:results}
\end{table}

\section{Experiments}
We performed three kinds of evaluation.
First, we evaluated the performance of the proposed ConvNet at classifying RC data at WSI level in cross-validation.
Second, we evaluated the performance of the same ConvNet, trained on RC data, applied to the independent CRC data set.
In this second experiment, we compared two state-of-the-art stain normalization algorithms by matching the color space of CRC data with the RC data used for training.
As part of the second experiment, we also analyze the importance of stain normalization applied to training and test data.

\begin{figure}[t]
\centering
{\includegraphics[width=1.\linewidth]{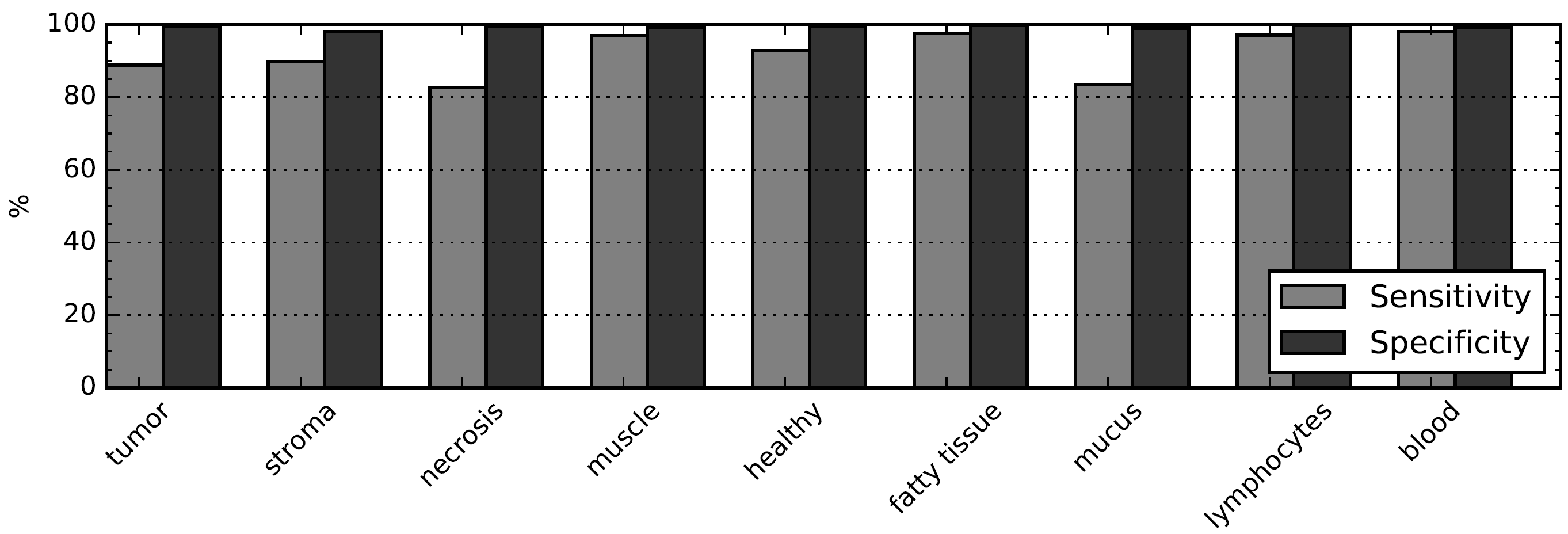}}\hspace{0.0em}\vspace{-0.5cm}
\caption{Per-class sensitivity and specificity of the proposed ConvNet on the 9-class rectal cancer dataset.\vspace{-0.3cm}
\label{fig:quantitative_results}}
\end{figure}

\subsection{Rectal cancer tissue classification}\label{sec:experiment1}
We evaluated the performance of the ConvNet on the RC dataset using a 5-fold cross-validation approach.
We build each fold selecting 40 WSI for training, 19 for validation and 15 for test, without overlap of patients across datasets.
For each fold, the ConvNet was trained with 4000 iterations of stochastic gradient descent with a constant learning rate of 0.0003, the ADAM algorithm for the update of the parameters and categorical cross-entropy as loss function.
For each iteration, a mini-batch of 256 patches was built on-the-fly by randomly sampling a balanced amount of patches from each manually annotated class in the training set.
Data was augmented by rotating each patch of 90, 180 and 270 degrees, increasing the amount of training patches to approximately 4 million per fold.
During training, the performance of the system was monitored  by classifying a fixed validation set of 45000 samples (5000 per class).
After training, the performance was measured by comparing the label of pixels in the regions manually annotated and the result of the classifier over the 74 WSIs in the RC dataset.
An overall 9-class accuracy of 93.8\% was obtained.
Overall performance of per-class sensitivity and specificity are reported in Figure \ref{fig:quantitative_results}. \vspace{-0.2cm}

\subsection{The importance of stain normalization} 
\begin{figure}[t]
\centering
\subfigure[\label{a}]
{\includegraphics[width=0.16\linewidth]{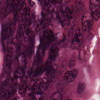}}
\subfigure[\label{b}]
{\includegraphics[width=0.16\linewidth]{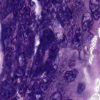}}
\subfigure[\label{c}]
{\includegraphics[width=0.16\linewidth]{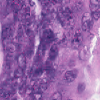}}\hspace{0.1em}
\subfigure[\label{d}]
{\includegraphics[width=0.16\linewidth]{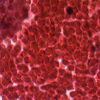}}
\subfigure[\label{e}]
{\includegraphics[width=0.16\linewidth]{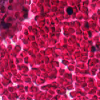}}
\subfigure[\label{f}]
{\includegraphics[width=0.16\linewidth]{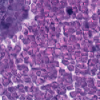}}\vspace{-0.5em}
\caption{Samples from CRC$_p$ for tumor (a) and debris (d) class, together with result of SN$_1$ (b, e) and of SN$_2$ (c, f).\vspace{-0.3cm}
\label{fig:comparison}}
\end{figure}

The definition of classes in our RC data is slightly different from the one coming with the CRC dataset.
Therefore, we grouped corresponding classes as described in Table \ref{tab:classes}, which resulted in a 6-class problem, used to evaluate the performance of the trained ConvNet applied to CRC data.
Applying the ConvNet directly to CRC data gave a poor accuracy value of 50.96\% (experiment A).
Based on this result, we investigated the role of staining in this classification task.
For this purpose, we selected a representative RC image from the RC cohort as the template image to define the mapping parameters $\theta$ for the SN$_1$ and SN$_2$ algorithms\footnote{The function $f(I,\theta)$ for SN$_1$ is available in the form of look-up table at https://github.com/francescociompi/stain-normalization-isbi-2017}.
Successively, we stain-normalized patches in the CRC dataset and classified it again with the same ConvNet.
Stain normalization allowed to remarkably improve the accuracy by more than 20\%, reaching values of 75.55\% (experiment B) and 73.99\% accuracy for SN$_1$ and SN$_2$ respectively.
Besides the difference in accuracy, we found that SN$_2$ failed to normalize 3 patches of adipose tissue, and it tends to force color normalization regardless of the type of tissue considered.
As an example, in Figure \ref{fig:comparison}, two CRC patches are depicted, containing tumor and blood cells respectively.
While both SN$_1$ and SN$_2$ correctly normalize stain for the tumor patch, the characteristic red color of blood cells is completely lost using SN$_2$, while SN$_1$ keeps a substantial component of the red channel.

\begin{figure}[t]
\centering
{\includegraphics[width=0.24\linewidth]{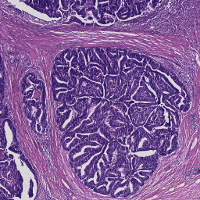}}\hspace{0.0em}
{\includegraphics[width=0.24\linewidth]{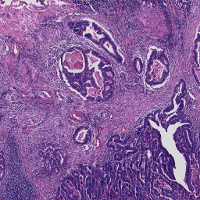}}\hspace{0.0em}
{\includegraphics[width=0.24\linewidth]{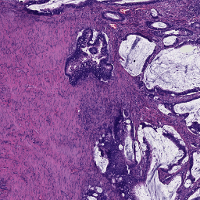}}\hspace{0.0em}
{\includegraphics[width=0.24\linewidth]{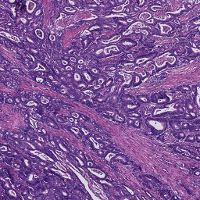}}\hspace{0.0em}
{\includegraphics[width=0.24\linewidth]{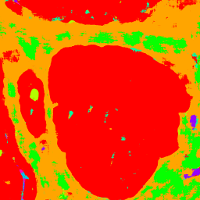}}\hspace{0.0em}
{\includegraphics[width=0.24\linewidth]{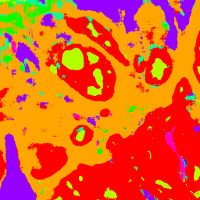}}\hspace{0.0em}
{\includegraphics[width=0.24\linewidth]{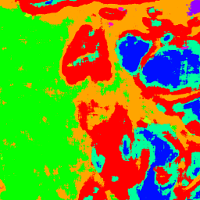}}\hspace{0.0em}
{\includegraphics[width=0.24\linewidth]{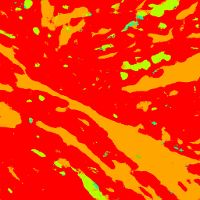}}\vspace{0.2em}
{\includegraphics[width=0.99\linewidth]{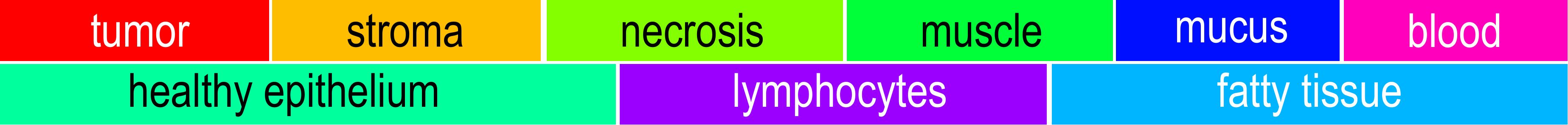}}\hspace{0.0em}
\caption{ConvNet tissue classification results on CRC$_t$ data.
\label{fig:qualitative_results}}
\end{figure}

Based on the positive result of the previous experiment, the question remains whether stain normalization should only be applied to test data or also to training data.
To address this question, we introduced SN$_1$ in our training and testing pipeline and evaluated the accuracy of a ConvNet trained on RC data and applied to CRC data in four different configurations, namely with (experiment C) and without (experiment D) stain normalization applied to RC (training) data and to CRC (test) data.
The results are reported in Table \ref{tab:results}, where for the sake of clarity a letter is assigned to the result of each combination.
As expected, experiment D resulted in the lowest performance, since the ConvNet does not learn any stain variability from the training set, while such a variability is expected in the test set.
Although all the training cases come from the same cohort and staining was done in the same lab, experiment C gave an improvement of $\approx$ 5\% accuracy compared to B, which indicates that the ConvNet can benefit from the variability in the training set.
It is worth noting that experimental settings of A represents what is commonly done in digital pathology research, where training data from a given cohort, with some variability are used to train a classifier, whose performance are evaluated on an independent set of data coming from different laboratories and stained with different procedures.
Experiment B reduced the variability in the test set by adapting data to match the stain distribution of the template image.
Finally, experiment C showed a substantial improvement compared to all other possible combinations of settings.
Qualitative results of images in CRC$_t$ classified under the settings of experiment C are depicted in Figure \ref{fig:qualitative_results}.\vspace{-0.3cm}

\section{Conclusion}\blfootnote{The authors would like to thank NVIDIA Corporation for the donation of a GeForce GTX Titan X graphics card used in the experiments. This project was supported by the Alpe dHuZes / Dutch Cancer Society Fund (grant number KUN 2014-7032).}
We have presented an approach based on convolutional networks for multi-class classification of CRC tissue in H\&E histopathology images.
Applying stain normalization to training and test data takes out of the equation most of the sources of variability due to staining.
Based on our experiments, we conclude that stain normalization is a necessary step to include in the training and evaluation pipeline of an automatic system for CRC tissue classification based on ConvNets.


%


\end{document}